\documentclass[conference]{IEEEtran}
\IEEEoverridecommandlockouts
\usepackage{cite}
\usepackage[english]{babel}
\usepackage{amsthm}
\usepackage{balance}

\theoremstyle{definition}

\usepackage{algorithm}
\usepackage{algpseudocode}

\algrenewcommand\algorithmicrequire{\textbf{Input:}}
\algrenewcommand\algorithmicensure{\textbf{Output:}}

\usepackage{amsmath}
\usepackage{amsfonts}
\usepackage{mathtools}
\usepackage{bm}

\newcommand{\IS}{\mathcal{X}} 
\newcommand{\HS}{\mathcal{H}} 

\DeclareMathOperator*{\argmax}{arg\,max}

\def\reals{\mathbb{R}}  

\usepackage{amsmath,amssymb,amsfonts}
\usepackage{graphicx}
\usepackage{textcomp}
\usepackage{xcolor}
\def\BibTeX{{\rm B\kern-.05em{\sc i\kern-.025em b}\kern-.08em
    T\kern-.1667em\lower.7ex\hbox{E}\kern-.125emX}}
\begin{document}

\title{Molecular Classification Using Hyperdimensional Graph Classification}

\author{\IEEEauthorblockN{Pere Vergés}
\IEEEauthorblockA{\textit{University of California Irvine} \\
Irvine, USA \\
pvergesb@uci.edu}
\and
\IEEEauthorblockN{Igor Nunes}
\IEEEauthorblockA{\textit{University of California Irvine} \\
Irvine, USA \\
igord@uci.edu}
\and
\IEEEauthorblockN{Mike Heddes}
\IEEEauthorblockA{\textit{University of California Irvine} \\
Irvine, USA \\
mheddes@uci.edu}
\and
\IEEEauthorblockN{Tony Givargis}
\IEEEauthorblockA{\textit{University of California Irvine} \\
Irvine, USA \\
givargis@uci.edu}
\and
\IEEEauthorblockN{Alexandru Nicolau}
\IEEEauthorblockA{\textit{University of California Irvine} \\
Irvine, USA \\
nicolau@ics.uci.edu}
}

\author{\IEEEauthorblockN{Pere Vergés,
Igor Nunes, Mike Heddes,
Tony Givargis and Alexandru Nicolau
}
\IEEEauthorblockA{University of California Irvine\\
pvergesb@uci.edu, igord@uci.edu, mheddes@uci.edu, givargis@uci.edu, nicolau@ics.uci.edu}}

\maketitle

\begin{abstract}
 Our work introduces an innovative approach to graph learning by leveraging Hyperdimensional Computing. Graphs serve as a widely embraced method for conveying information, and their utilization in learning has gained significant attention. This is notable in the field of chemoinformatics, where learning from graph representations plays a pivotal role. An important application within this domain involves the identification of cancerous cells across diverse molecular structures.
 We propose an HDC-based model that demonstrates comparable Area Under the Curve results when compared to state-of-the-art models like Graph Neural Networks (GNNs) or the Weisfieler-Lehman graph kernel (WL). Moreover, it outperforms previously proposed hyperdimensional computing graph learning methods. Furthermore, it achieves noteworthy speed enhancements, boasting a 40x acceleration in the training phase and a 15x improvement in inference time compared to GNN and WL models. This not only underscores the efficacy of the HDC-based method, but also highlights its potential for expedited and resource-efficient graph learning.
\end{abstract}


\section{Introduction}
\begin{figure}[htbp!]
\centering
\includegraphics[width=5.6cm]{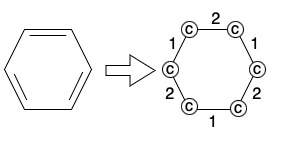}
\caption{\label{fig:benzene} Benzene to graph representation~\cite{4812459}.}
\end{figure}

Molecular classification is a crucial task in pharmaceutical research~\cite{B409813G,doi:10.1021/jm0582165}, gaining increased prominence in recent years. Various techniques, including statistical models, decision trees~\cite{doi:10.1021/ci9903049,ZMUIDINAVICIUS2003621}, Bayesian models~\cite{doi:10.1021/ci7003253,doi:10.1142/9789814447300_0044}, kernel-based methods~\cite{Frhlich2005OptimalAK, PMID:11604029, doi:10.1021/ci049737o}, neural networks~\cite{doi:10.1021/ci9903049,PMID:11121522}, and sub-graph mining techniques~\cite{10.1145/1376616.1376662,4812459}, are employed in this field. The quantitative structure-activity relationship (QSAR)\cite{doi:10.1142/9789814447300_0044} serves as the foundation for these techniques, leveraging the assumption that molecular structure reflects multiple chemical properties. In Figure~\ref{fig:benzene}, we show an example of the conversion of benzene to a graph. Analyzing the graph structure of molecules provides valuable insights into their relationships and similarities.

Among the molecular classification approaches, methods based on feature vector representation have emerged, particularly utilizing chemical descriptors and fingerprints. Fingerprints~\cite{10.1145/1459352.1459353,ADAMSON1973561,fingerprint,fingerprint2}, for instance, generate fixed-width bit-vectors by enumerating cycles and linear paths, offering a comprehensive representation of molecular features.

Another set of approaches involves mining frequent substructures based on biological and chemical properties of molecules~\cite{Mrzic2018GraspingFS,article}. Such methods generate feature vectors indicating the presence of distinct properties, requiring a frequency threshold to differentiate between frequent and infrequent substructures. Notably, studies suggest that infrequent substructures are more effective in discriminating between different molecule types~\cite{4812459}.

This work integrates the insights of infrequent substructures learned from graph mining works~\cite{4812459} with Hyperdimensional Computing (HDC)\cite{kanerva2009hyperdimensional}. As Machine Learning gains popularity for solving complex problems\cite{AlphaTensor2022}, HDC has emerged as a favored solution for resource-constrained scenarios~\cite{vergés2023hdcc,8885251}, spanning voice and gesture recognition~\cite{rahimi2016hyperdimensional,imani2017voicehd}, natural language processing~\cite{rahimi2016robust}, graph learning~\cite{nunes2022graphhd,graphd}, and bioinformatics~\cite{9116397}. HDC appeal lies in its ability to abstract information efficiently.

Beyond HDC, there is a growing interest in graph learning techniques, specifically those focused on graph similarity~\cite{Morris+2020}. Graph kernels, commonly coupled with support vector machines (SVM)\cite{Keerthi2006BuildingSV}, have gained attention. Nevertheless, their computational cost scales quadratically, rendering them unsuitable for online learning and real-time applications\cite{10.1007/978-3-540-30497-5}. Another popular approach is the use of Graph Neural Networks (GNN). However, GNNs present challenges due to their computational time complexity and energy costs, thereby limiting their applicability in real-world scenarios, such as embedded or IoT applications~\cite{6424332,8567692,10296393} that are subject to strict design constraints.

Our research introduces a novel method for detecting cancerous cells by comparing the cell structures of cancerous and non-cancerous cells and employing graph classification based on HDC. The approach is based on sub-graph kernel matching~\cite{10.5555/3042573.3042614}, taking into account infrequent cell patterns to refine the learned associative memory using HDC adaptive learning techniques. The primary objective of our method is to improve efficiency while ensuring a level of accuracy comparable to state-of-the-art graph learning methods in the context of cancerous cell detection. The applicability of our work extends to various real-time scenarios requiring online learning for timely execution~\cite{4812459,JMLR:v12:shervashidze11a,DBLP:journals/corr/abs-1901-00596}, such as the evaluation of evolving cells in human liver diseases~\cite{cells9020386} or gliomas~\cite{KEUNEN201498}; and in anomaly detection for molecular data~\cite{luo2022deep,10.1093/bioinformatics/btad501}, where approaches like GNNs or Weisfeiler-Lehman (WL) kernels may prove too expensive for real-time conditions.


\section{Related Work}
In this section, we present related works on graph learning methods, covering techniques for molecular classification. 

\subsection{Graph Learning Methods}
\paragraph{Kernel methods} are popular machine learning techniques for comparing data points within a graph using similarity metrics. The literature encompasses various kernel methods, each focusing on different properties of a graph. Notable examples include the Random Walk Kernel~\cite{10.5555/2969239.2969422}, Tree Pattern Kernel~\cite{tree_pattern}, Shortest-Path Kernel~\cite{nikolentzos-etal-2017-shortest}, Optimal Assignment Kernel~\cite{inbook}, Graphlet Kernel~\cite{pmlr-v5-shervashidze09a}, Weisfeiler-Lehman Kernel~\cite{JMLR:v12:shervashidze11a}, Subgraph Matching Kernel~\cite{4812459}, among others. From this extensive set, we will focus on comparing our results with the state-of-the-art Weisfeiler-Lehman (WL) kernel.

The WL kernel iteratively refines label assignments for graph nodes based on their neighbors' structure. The iterative algorithm updates node labels until no further changes occur. The execution cost depends on the neighborhood's number of hops, increasing with larger values. An enhancement to the WL algorithm involves combining it with Optimal Assignment~\cite{10.1145/1102351.1102380}, determining node pairs with minimal cost in graph comparisons according to a similarity metric.

\paragraph{Graph Neural Network (GNN)} Recent advances in Graph Neural Networks (GNNs) involve iteratively updating representations by exchanging information with their neighbors, and once trained, they are used to perform classification. Despite the popularity of GNNs, graph kernels remain competitive. In fact, it has been shown that GNNs are, at most, as powerful as the Weisfeiler-Lehman (WL) test in graph discernment \cite{xu2019powerful}.

\paragraph{Fingerprints} These are well-known techniques in cheminformatics for representing molecules using feature vectors~\cite{fingerprint,fingerprint2}. These fingerprints are generated by enumerating all molecular graph substructures and extracting predefined relevant substructures. Feature vectors can then be compressed using hashing techniques. In order to perform classification using fingerprints one can use similarity metrics to define graph similarity. This is typically done using the Tanimoto coefficient~\cite{tanimoto}.

\paragraph{Subgraph mining} This is another prevalent approach applicable to large graphs. Various techniques focus on mining frequent patterns, with GraphSig~\cite{4812459} yielding superior results. GraphSig mines significant subgraphs, translating them into the feature vector space by evaluating statistical significance patterns, which helps to use the most relevant features of the graph when performing graph classification, which is performed using similarity measures. This technique highlights that infrequent patterns are more significant than frequent ones. 

\section{Hyperdimensional Computing (HDC)}
In this section, we provide a brief introduction to HDC, explain how classification is performed, and introduce two graph encoding approaches proposed in the literature.

\subsection{Information Representation}\label{sec:information_representation}
In HDC, information is encoded as points in a high-dimensional space (\textit{hyperspace}) $\HS$, known as \textit{hypervectors}. Given input data $x \in \IS$, a mapping $\phi \colon \IS \to \HS$ produces the \textit{encoding} $\phi(x)$. The core concept is that similar inputs in the original space map to similar vectors in the hyperspace.

\subsection{Operations}
\label{sec:operations}

Once information is in the hyperspace, operations like \textit{superposition}, \textit{binding}, and \textit{permutation} are applied to create composite representations. These operations preserve hypervector dimensionality, generating a hypervector in the same hyperspace as the operands. Their composability allows for versatile encoding tailored to diverse applications, capturing data compositionality effectively. The \textit{similarity} metric evaluates hypervector relations and resemblance.

\paragraph{Similarity metric}\label{sec:similarity} The similarity between hypervectors is given by $\delta: \HS\times\HS\to\reals$, and is used to determine relations, crucial for classification problems~\cite{olver2006applied}. Usually, the similarity metric is implemented using cosine similarity, dot product, or hamming distance.

\paragraph{Superposition}
The superposition operation aggregates information into a single hypervector representation using element-wise vector addition. This operation produces a vector maximizing similarity to the original operands. In certain hyperspaces, an inverse operation to bundling exists, removing an item from the bundled representation by applying superposition with the additive inverse~\cite{plate2003holographic, kanerva1988sparse}.

\paragraph{Binding}
The binding operation associates two hypervectors, yielding a hypervector dissimilar to both inputs. Widely employed for information association, similar to assigning values to variables, binding has an inverse operation (unbinding) involving the multiplicative inverse. Depending on the hyperspace, the binding operation can be self-inverse or may require a different operation~\cite{plate2003holographic, kanerva1988sparse}.

\paragraph{Permutation}
The permutation operator is used for imposing order on hypervectors. This operation yields a dissimilar hypervector, enabling the assignment of specific orders within the hyperspace. Importantly, the inverse operation allows exact retrieval of the original input hypervector, ensuring reversibility in the permutation process~\cite{cohen-widdows-2018-bringing}.

\subsection{Hyperdimensional Classification}\label{section:classification}
Hyperdimensional classification~\cite{HDC_classification} comprises two stages. Firstly, the training stage involves learning from the provided input data. Secondly, the inference stage involves the model predicting the class of unseen samples.

\paragraph{Training}
The initial training step involves encoding input data samples into hypervectors by employing the HDC operations. 
Subsequently, the encoded sample is added to the associative memory, which is comprised of hypervectors that represent the prototypes of the different classes. This addition entails element-wise addition of the encoded sample with the corresponding class hypervector, identified by its label.

\begin{figure}[htbp!]
\centering
\includegraphics[width=5.6cm]{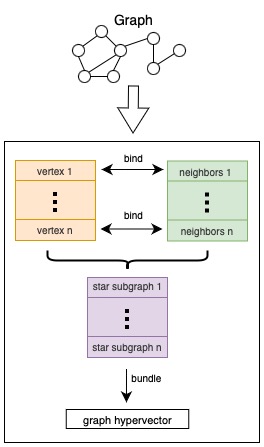}
 \caption{\label{fig:encoding} Proposed graph encoding.}
\end{figure}

\paragraph{Inference}
The inference process consists of three steps. Similar to training, the initial inference step involves encoding input data into the hyperdimensional space. Ensuring consistency and accuracy, it is crucial to use the same encoding for both phases. Following encoding, the next step computes the similarity between the encoded sample and associative memory. This yields a list of distance measures between the sample and different classes. Finally, the class most similar to the encoded sample is selected from the list of similarities for predicting the input data class.

\subsection{Hyperdimensional graph encodings}
The first work that proposed encoding graphs into hyperspace is due to Gayler and Levy~\cite{Gayler2009ADB}.  Poduval et al.~\cite{graphd} built up on it and demonstrated different properties of the encoding. Another work, named GraphHD~\cite{nunes2022graphhd}, proposed creating an ordering of the nodes using a node centrality metric when assigning each one its hypervector for classification. This work shows that using HDC on certain datasets gives comparable results to state-of-the-art models such as GNNs or WL kernels, while having faster training and inference times.

Our method uses hyperdimensional classification method for encoding the graph into hypervectors. The approach takes on the insights that are learnt about infrequent patterns from subgraph mining techniques, by creating the encoding representation using star subgraphs and giving more importance to the graph infrequent patterns by using RefineHD~\cite{EasyChair:11191}. This allows our method to have similar execution times to HDC approaches while achieving comparable accuracy results to the state-of-the-art graph learning methods.

Our approach employs a hyperdimensional classification method, which involves encoding the graph into hypervectors. By leveraging the knowledge gained from subgraph mining techniques, particularly focusing on infrequent patterns, and subgraph matching kernels, our method employs star subgraphs in the encoding representation. To emphasize the importance of graph infrequent patterns, we utilize RefineHD~\cite{EasyChair:11191}. This strategy ensures that our approach maintains execution times similar to HDC approaches, while also delivering accuracy results comparable to state-of-the-art graph learning methods.

\section{Our Method}
The encoding process in HDC poses a significant challenge and holds paramount importance as it dictates how data is captured. Our primary focus in this study is on defining a method to encode the graph of cells effectively, with the goal of capturing maximal information. This capability is crucial for distinguishing whether a molecule exhibits an active or inactive cell line. Consequently, our input comprises structural graphs, and we must subsequently map this information into the hyperdimensional space.


\begin{table}[]
 \caption{\label{table:centrality metric} Centrality metrics AUC performance using GraphHD.}
\centering
\begin{tabular}{|l|l|}
\hline
Centrality metric      & AUC \\ \hline
None (Gayler \& Levy)               & \textbf{61.12}    \\ \hline
PageRank (GraphHD)             & 59.57    \\ \hline
Degree centrality      & 59.70    \\ \hline
Closeness centrality   & 58.99    \\ \hline
Betweenness centrality & 59.57    \\ \hline
Load centrality        & 59.84    \\ \hline
Subgraph centrality    & 59.38    \\ \hline
Harmonic centrality    & 58.83    \\ \hline
\end{tabular}
\end{table}


Before introducing our method, we conducted an evaluation of centrality metrics to be applied to the GraphHD encoding, since their work showed great results when using PageRank centrality metric~\cite{Page1999ThePC}. Our findings indicate that centrality metrics are not effective in capturing the graph's information when applied to HDC and molecule graphs. Table~\ref{table:centrality metric} illustrates the average AUC (Area Under the Curve), on the PTC\_FM, MUTAG, NCI1, ENZYMES, and PROTEINS datasets, of GraphHD using different centrality metrics. Notably, the first scenario, employing random hypervectors without considering centrality metrics, outperforms all cases where nodes are ordered using such metrics. This observation underscores the limited efficacy of centrality metrics in enhancing the AUC of the encoding process.

\begin{algorithm}
\begin{algorithmic}
\caption{Graph classification algorithm: Training}\label{alg:training}
\Require{Training set of graphs and their class labels $\mathcal{G}$}
\Ensure{Memory $\mathcal{M}$ containing the set of class hypervectors $C_1 $, \dots , $C_k$}
\For{$(g, i) \in \mathcal{G}$}
\State $H_g \gets \emptyset$
\For{node $v \in V(g)$}
\State $H_v \gets \phi(v)$
\For{$u \in \mathrm{neighbours}(v)$}
\State $H_v \gets \mathrm{bind}(H_v, \phi(u))$
\EndFor
\State $H_g \gets \mathrm{bundle}(H_g, H_v)$
\EndFor
\State $\mathcal{M} \gets \mathrm{RefineHD}(H_g, i, \mathcal{M})$
\EndFor
\end{algorithmic}
\end{algorithm}

Our approach is inspired by subgraph-kernel matching. One approach that uses subgraph matching kernel~\cite{10.5555/3042573.3042614} counts the number of matching subgraphs between two graphs. Taking this idea into account, we generated a graph encoding that creates a histogram of star subgraphs. Figure \ref{fig:encoding} depicts the proposed graph encoding.

By performing a histogram of star subgraphs, we make our representation more complex compared to the previously proposed methods. This helps to make a more drastic distinction between different graph patterns. Moreover, our method builds on the idea shown in~\cite{4812459}, where they found that when classifying molecules using their structure, in this study~\cite{4812459} was shown that subgraphs that appear in the graph with low frequency carry more information than frequent graphs. For this reason, we decided to apply adaptive learning to our classification algorithm. This will help in giving less weight to frequent graphs and subgraphs and giving more value to those that are more infrequent.

\begin{table}[hbtp!]
\caption{\label{table:adaptive} Adaptive learning AUC.}
\centering
\begin{tabular}{|l|l|l|l|l|}
\hline
Model    & Add   & AdaptHD & OnlineHD & RefineHD         \\ \hline
AUC      & 76.32 & 89.36   & 89.41    & \textbf{90.51} \\ \hline
\end{tabular}
\end{table}

In HDC, adaptive learning is a common technique to improve AUC while maintaining good efficiency, although this technique can also be applied to perform iterative training, which augments its training time considerably. There are three proposed models: AdaptHD~\cite{8918974}, OnlineHD~\cite{9474107}, and RefineHD~\cite{EasyChair:11191}. We have evaluated all these models, and we have seen that RefineHD is the one that gives the best AUC for this application. In Table~\ref{table:adaptive}, we show the comparison between the different methods of adding information to the associative memory. The first method (Add) adds all graph encodings to the memory. The second method (AdaptHD) only adds misclassified patterns to the memory. The third method (OnlineHD) only adds misclassified patterns but in a weighted manner, taking into account the cosine similarity of the sample and the class. The last method (RefineHD) adds both classified and misclassified samples in a weighted manner but only adds a sample when the cosine similarity to the correct class is below a threshold, which is partially computed using the similarities of the incorrectly classified samples. From these methods, we see that RefineHD is the one that behaves the best, having a +1\% improvement compared to the other adaptive learning methods.

Since RefineHD was the method that behaved the best, we decided to conduct a study and see what threshold performed better. In Table~\ref{table:threshold_value}, we see in the results that having a threshold value between 1.5 and 2 gives the best results.

\begin{table}[]
\caption{\label{table:threshold_value} RefineHD threshold value.}
\centering
\begin{tabular}{|l|l|l|l|l|}
\hline
Threshold    & 1  & 1.5 & 1.8 & 2         \\ \hline
AUC      & 90.51 & 91.00   & \textbf{91.09}   & 90.98 \\ \hline
\end{tabular}
\end{table}

We show the training pseudocode of our graph classification in Algorithm~\ref{alg:training}, for each class we will encode all graphs using the proposed star subgraph encoding that uses bind as the HDC to build the subgraphs, in order to add a graph to its class we use the RefineHD method.

\begin{algorithm}
\caption{Graph classification algorithm: Inference}\label{alg:infrence}
\begin{algorithmic}
\Require{Graph $g$ and trained memory $\mathcal{M}$ containing the set of class hypervectors $C_1 $, \dots , $C_k$}
\Ensure{Predicted label $\hat{y}$}
\State $H_g \gets \emptyset$
\For{node $v \in V(g)$}
\State $H_v \gets \phi(v)$
\For{$u \in \mathrm{neighbours}(n)$}
\State $H_v \gets \mathrm{bind}(H_v, \phi(u))$
\EndFor
\State $H_g \gets \mathrm{bundle}(H_g, H_v)$
\EndFor
\State $\hat{y} \gets  \argmax_i(\delta(H_g, C_i))$
\end{algorithmic}
\end{algorithm}

For inference, it is necessary to encode the graph using the hyperdimensional vectors that represent each node. After encoding, perform the cosine similarity operation against the previously trained memory. This will return a value between 0 and 1 for each class. The resulting prediction will be the one that has the value closest to one, indicating higher similarity to the corresponding class.

\section{Experiments}
To evaluate our method, we utilized eleven anticancer screen datasets from the TUDataset~\cite{Morris+2020}, which were extracted from PubChem\footnote{https://pubchem.ncbi.nlm.nih.gov}. The method was implemented using TorchHD~\cite{JMLR:v24:23-0300}, a Python library dedicated to HDC. We compared the performance of our model against both previous HDC graph classification methods and state-of-the-art graph classification methods. We chose to compare it against the 1-WL sparse graph kernel, as the dense 1-WL and WL-OA are too computationally expensive for cancer cells detection. Additionally, we conducted a comparison with a GNN model based on the Graph Isomorphism Network (GIN)~\cite{gin}. The configuration of these two methods adhered to the default settings from the benchmark evaluation of the TUDataset. All experiments were executed on a machine with 2 x 6-core Intel Xeon X5680 @ 3.33GHz, 96GB RAM, 64-bit CentOS Linux 7. For each experiment, we utilized 10,000 dimensions and conducted 10 repetitions.

\subsection{Dataset}
The datasets were extracted from PubChem, which contains a compilation of biological activities of different molecules. We utilized the records of anticancer screen data tested against various cancer cell lines. Each of these datasets contains molecules that have been tested against cancer cells, along with the results indicating whether they are active or inactive. A summary of the datasets we have used is presented in Table~\ref{table:datasets}.

\begin{table}[]
\caption{\label{table:datasets} Anticancer screen datasets.}
\centering
\begin{tabular}{|l|l|l|}
\hline
Dataset  & Size  & Description            \\ \hline
MCF-7    & 28972 & Breast                 \\ \hline
MOLT-4   & 41810 & Leukemia               \\ \hline
NCI-H23  & 42164 & Non-small Cell lung    \\ \hline
OVCAR-8  & 42386 & Ovarian                \\ \hline
PC-3     & 28679 & Prostate               \\ \hline
P388     & 46440 & Leukemia               \\ \hline
SF-295   & 40350 & Central nervous system \\ \hline
SN12C    & 41855 & Renal                  \\ \hline
SW-620   & 42405 & Colon                  \\ \hline
UACC-257 & 41864 & Melanoma               \\ \hline
Yeast    & 83933 & Yeast anticancer       \\ \hline
\end{tabular}
\end{table}

\begin{table}[hbtp!]
\centering
\caption{\label{table:adaptive_learning} Adaptive learning AUC results.}

\begin{tabular}{|l|l|l|l|l|}
\hline
Dataset & Baseline   & AdaptHD & OnlineHD & RefineHD       \\ \hline
MCF-7   & 54.64 & 86.81   & 86.63    & \textbf{88.43} \\ \hline
MOLT-4  & 54.15 & 87.25   & 86.57    & \textbf{88.24} \\ \hline
NCI-H23 & 57.74 & 91.20   & 91.73    & \textbf{92.64} \\ \hline
OVCAR-8 & 56.88 & 91.40   & 91.44    & \textbf{92.44} \\ \hline
PC-3    & 56.49 & 90.33   & 90.44    & \textbf{91.41} \\ \hline
P388    & 56.52 & 91.00   & 91.14    & \textbf{92.14} \\ \hline
SF-295  & 57.97 & 91.70   & 91.81    & \textbf{92.27} \\ \hline
SN12C   & 58.01 & 91.77   & 91.82    & \textbf{92.70} \\ \hline
SW-620  & 55.87 & 90.10   & 89.67    & \textbf{91.27} \\ \hline
UACC257 & 57.00 & 92.60   & 92.70    & \textbf{93.78} \\ \hline
Yeast   & 51.14 & 78.84   & 79.53    & \textbf{80.33} \\ \hline
\end{tabular}
\end{table}

\subsection{Adaptive learning}
The first experiment we conducted aimed to determine which adaptive learning algorithm would perform best for detecting cancerous cells. As explained in Section~\ref{section:classification}, we compared the hyperdimensional classification baseline to AdaptHD, OnlineHD, and RefineHD. As shown in Table~\ref{table:adaptive_learning}, RefineHD performed the best in terms of AUC results for all datasets. In Table~\ref{table:adaptive_time}, we present the average time per sample for both training and inference. Despite RefineHD being slower compared to the other methods, we chose it for our method since it yielded the best performance across all other methods.

\begin{table}[hbtp!]
\centering
\caption{\label{table:adaptive_time} Adaptive learning techniques train and inference times per sample.}
\begin{tabular}{|l|l|l|l|l|}
\hline
VSA        & Baseline & AdaptHD & OnlineHD & RefineHD \\ \hline
Train time & 0.42     & 0.57    & 0.58     & 0.62    \\ \hline
Inference time  & 0.53     & 0.65    & 0.64     & 0.69    \\ \hline
\end{tabular}
\end{table}

\subsection{Vector Symbolic Architectures}
The second experiment we conducted involved evaluating our proposed graph encoding using RefineHD and trying different Vector Symbolic Architectures (VSA)~\cite{gayler2003vector}. We experimented with Multiply Add Permute (MAP), Fourier Holographic Reduced Representation (FHRR), and Vector-Derived Transformation Binding (VTB).

Table~\ref{table:vsa} shows the performance results with the three different VSA, and we observe that in terms of AUC, the resulting values are very similar. However, in terms of execution time, the MAP architecture is the most efficient one.

\begin{table}[hbtp!]
\centering
\caption{\label{table:vsa} Vector Symbolic Architectures performance.}
\begin{tabular}{|l|l|l|l|}
\hline
VSA        & MAP   & FHRR  & VTB   \\ \hline
AUC        & 90.51 & 90.78 & 90.33 \\ \hline
Train time & 0.82  & 0.92  & 5.36  \\ \hline
Inference time  & 0.87  & 0.97  & 5.38  \\ \hline
\end{tabular}
\end{table}

\subsection{Methods comparison}
The third experiment involved comparing our method to both hyperdimensional graph classification methods (GraphHD and GraphHD) and state-of-the-art graph models (1-WL sparse and GIN). We attempted to execute against dense 1-WL and WLOA; however, the execution time was too large to be applied to cancer cell detection, since the execution of just one dataset was taking several days.

\begin{figure}[htbp!]
\centering
\includegraphics[width=8.6cm]{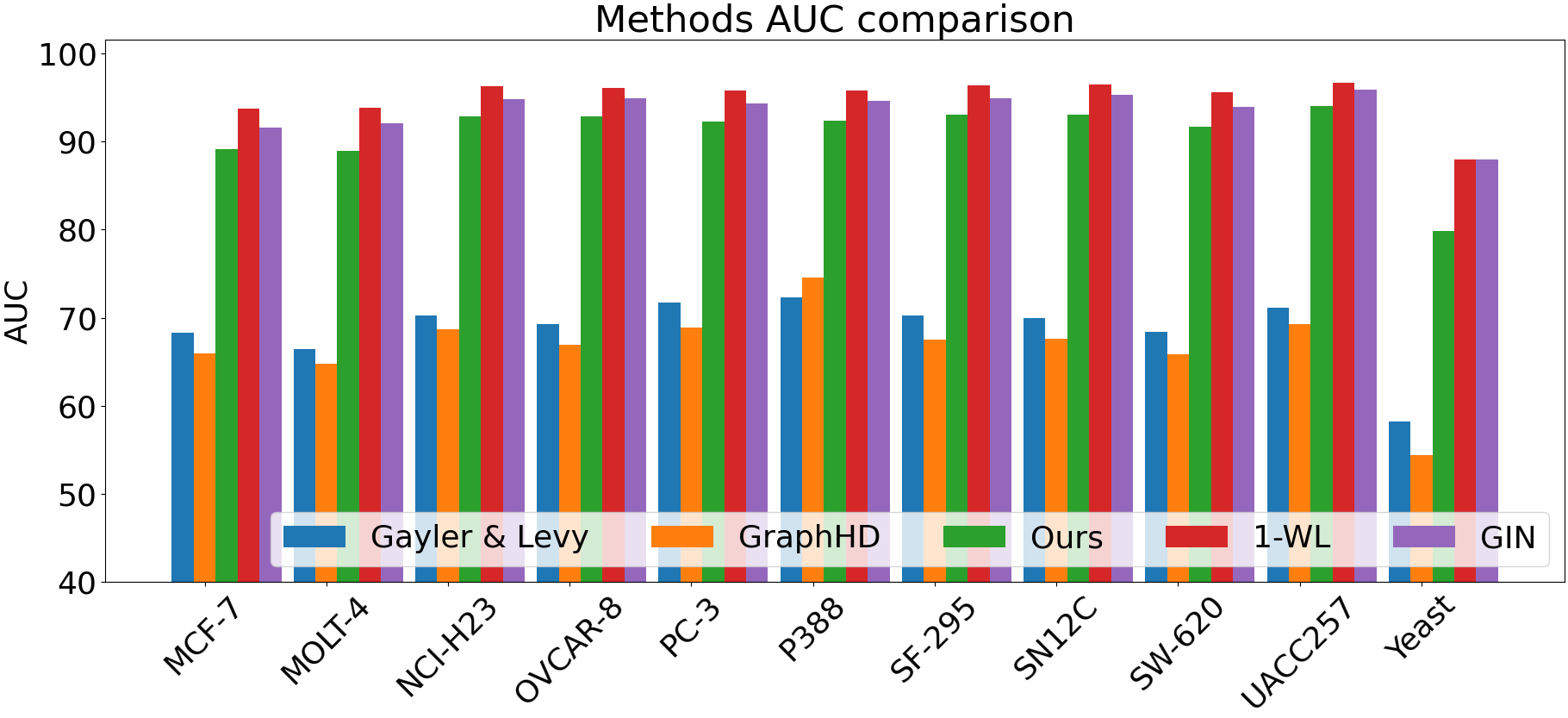}
 \caption{\label{fig:auc_methods} Method comparison techniques AUC.}
\end{figure}

In Figure~\ref{fig:auc_methods}, we observe the AUC of all the methods. As we can see, the two hyperdimensional classification models have considerably lower AUC values, lower by 20\% and 25\% compared to our proposed method. Compared to the state-of-the-art models, we have comparable results, lower by 3\% and 4\%. In terms of efficiency, our method is 1.25 times faster than (Gayler \& Levy encoding) and 1.33 times slower than GraphHD during training, with similar times for inference. Compared to the state-of-the-art models, we achieve large speedups in terms of training. Against WL, we get a 31x speedup, and a 41x speedup compared to GIN. In terms of inference, we have a speedup of 15x compared to WL and a slight speedup of 1.07x against GIN. All these results are shown in Tables~\ref{table:methods_train_time} and~\ref{table:methods_test_time}.

\begin{table}[hbtp!]
\centering
\caption{\label{table:methods_train_time} Method comparison train time.}
\begin{tabular}{|l|l|l|l|l|l|}
\hline
        & Gayler \& Levy & GraphHD & Ours & 1-WL  & GIN   \\ \hline
MCF-7   & 0.18   & 0.10    & 0.13 & 3.73  & 5.67  \\ \hline
MOLT-4  & 1.30   & 0.72    & 0.95 & 32.28 & 40.74 \\ \hline
NCI-H23 & 1.75   & 0.96    & 1.27 & 42.44 & 55.18 \\ \hline
OVCAR-8 & 1.15   & 0.63    & 0.84 & 26.80 & 36.43 \\ \hline
PC-3    & 1.21   & 0.66    & 0.89 & 23.85 & 37.15 \\ \hline
P388    & 0.61   & 0.39    & 0.46 & 15.25 & 12.80 \\ \hline
SF-295  & 1.36   & 0.75    & 0.99 & 31.94 & 45.72 \\ \hline
SN12C   & 1.18   & 0.65    & 0.87 & 27.79 & 37.76 \\ \hline
SW-620  & 0.12   & 0.06    & 0.09 & 2.95  & 3.93     \\ \hline
UACC257 & 0.01   & 0.01    & 0.01 & 0.44  & 0.60 \\ \hline
Yeast   & 0.03   & 0.02    & 0.02 & 1.25  & 0.66  \\ \hline
\end{tabular}
\end{table}

\begin{table}[hbtp!]
\centering
\caption{\label{table:methods_test_time} Method comparison inference time.}
\begin{tabular}{|l|l|l|l|l|l|}
\hline
        & Gayler \& Levy & GraphHD & Ours & 1-WL  & GIN  \\ \hline
MCF-7   & 0.20   & 0.11    & 0.14 & 1.65  & 0.18 \\ \hline
MOLT-4  & 1.45   & 0.83    & 1.01 & 15.95 & 1.12 \\ \hline
NCI-H23 & 1.94   & 1.12    & 1.36 & 21.60 & 1.39 \\ \hline
OVCAR-8 & 1.28   & 0.73    & 0.89 & 14.19 & 0.91 \\ \hline
PC-3    & 1.35   & 0.77    & 0.95 & 10.49 & 1.31 \\ \hline
P388    & 0.70   & 0.45    & 0.50 & 7.9   & 0.51 \\ \hline
SF-295  & 1.51   & 0.87    & 1.06 & 16.76 & 1.09 \\ \hline
SN12C   & 1.33   & 0.76    & 0.93 & 14.54 & 0.95 \\ \hline
SW-620  & 0.13   & 0.07    & 0.09 & 1.54  & 0.09  \\ \hline
UACC257 & 0.02   & 0.01    & 0.01 & 0.23  & 0.02 \\ \hline
Yeast   & 0.03   & 0.02    & 0.02 & 0.74  &  0.02 \\ \hline
\end{tabular}
\end{table}


\begin{table}[htbp!]
 \caption{\label{table:average results} Average AUC and train/inference times comparison.}
 \centering
\begin{tabular}{|l|l|l|l|}
\hline
Methods & AUC      & Train time   & Inference time    \\ \hline
Gayler \& Levy  & 68.73 (x0.75)         & 0.81 (x1.25)          & 0.91 (x1.42)         \\ \hline
GraphHD & 66.76 (x0.73)         & 0.45 (x0.75)          & 0.53 (x0.82)         \\ \hline
Ours    & 90.93 -          & 0.60 -                & 0.64 -          \\ \hline
WL      & 94.97 (x1.04)         & 18.98 (x31.6)        & 9.60 (x15)         \\ \hline
GIN      & 93.68 (x1.03)        & 25.15 (x41.91)       & 0.69 (x1.07)         \\ \hline
\end{tabular}
\end{table}

\subsection{Scalability}
\begin{figure}[htbp!]
\centering
\includegraphics[width=9cm]{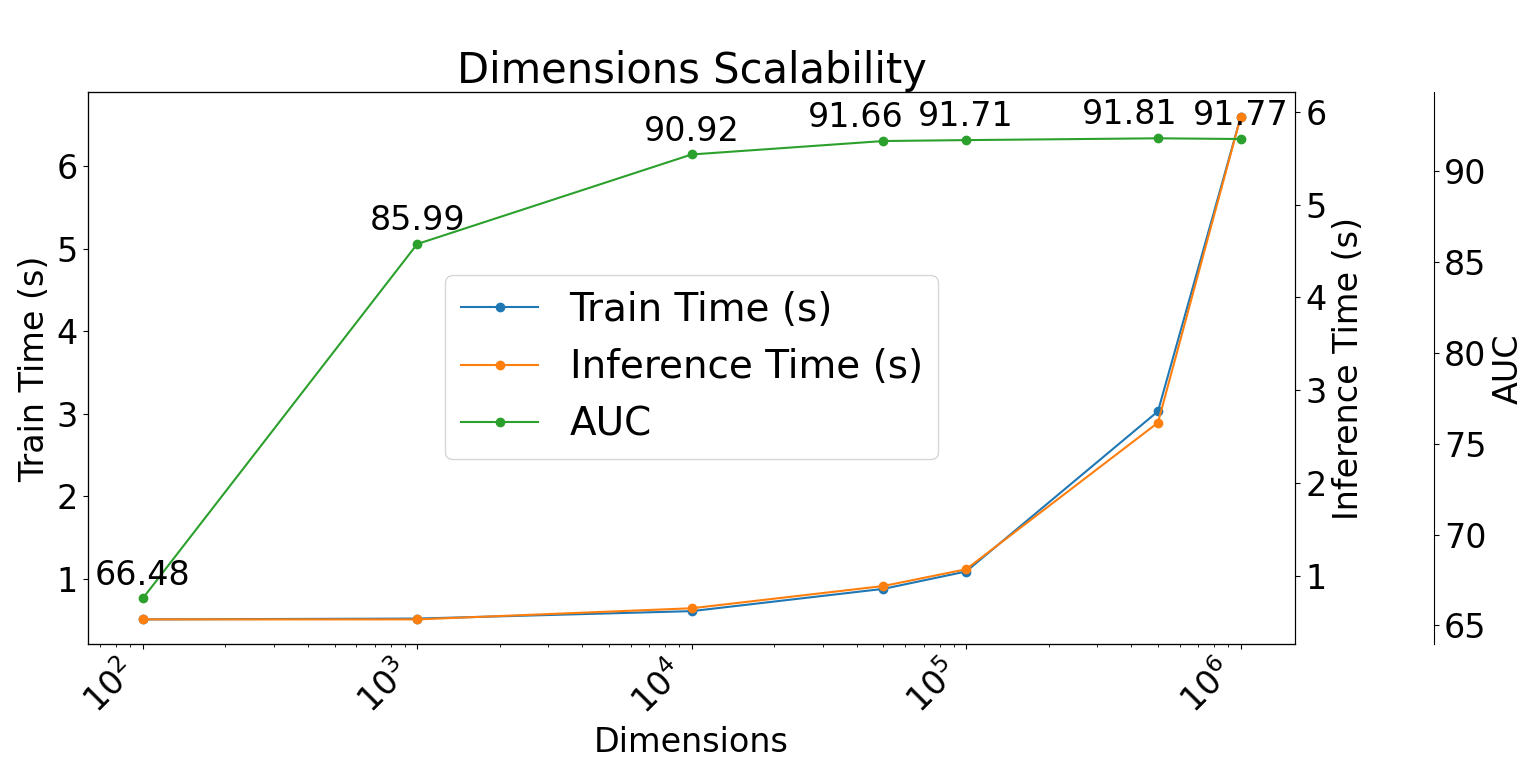}
 \caption{\label{fig:scalability} Method scalability.}
\end{figure}

Finally, the last experiment consisted of evaluating the scalability performance of this method. It aimed to assess both the AUC improvement achievable by adding more dimensions and the effect on training time and inference time. In Figure~\ref{fig:scalability}, we can observe that adding dimensions up to 50,000 results in an improvement in AUC. However, going beyond that range does not provide further improvement, as the AUC plateaus. Additionally, the training and inference times increase linearly with the number of dimensions.

\section{Conclusions}
This work introduces a novel graph encoding using HDC, employing star-subgraphs to represent graph information in hyperspace, and utilizing adaptive learning through the RefineHD approach to prioratize infrequent subgraph patterns. The focus is on representing graphs of molecules where a cancer cell is either active or inactive. The comparison with two graph encodings proposed in the literature using HDC demonstrates an improvement in accuracy of up to 25\% while maintaining a similar execution time. Additionally, our method is compared to state-of-the-art graph learning approaches such as Graph Neural Networks (GNNs) and the Weisfeiler-Lehman kernel, achieving comparable accuracy results and significant speedups of 40x during training and 15x during inference.
\balance
\bibliography{conference}

\end{document}